\documentclass[conference,letterpaper]{IEEEtran}

\usepackage[T1]{fontenc}
\usepackage[utf8]{inputenc}
\usepackage{microtype}
\usepackage{booktabs}
\usepackage{array}
\usepackage[table]{xcolor}
\usepackage{enumitem}
\usepackage[hidelinks,
            pdftitle={Towards an Agentic Operating System},
            pdfauthor={Gosia Steinder; Hubertus Franke}]{hyperref}
\usepackage{url}
\usepackage{cite}
\usepackage{graphicx}
\usepackage{hyphenat}
\usepackage{xspace}

\definecolor{ibmblue}{RGB}{15, 98, 254}

\setlist[itemize]{noitemsep, topsep=1pt}
\setlist[enumerate]{noitemsep, topsep=1pt}

\title{Towards an Agent Operating System -\\Lessons from Classical and Cloud OS}

\author{
\IEEEauthorblockN{Gosia Steinder, Hubertus Franke}
\IEEEauthorblockA{
IBM T.J.Watson Research Center\\
Yorktown Heights, NY 10598, USA\\
Email: \{steinder,frankeh\}@us.ibm.com}
}

\newcommand{\AOS}{Agent-OS\xspace}

\begin{document}
\maketitle

% ── Abstract ──────────────────────────────────────────────────────────────────
\begin{abstract}
Every major wave of platform software follows the same arc: an initial period
of experimentation with competing frameworks and ad-hoc implementations, followed
by the articulation of a small set of stable abstractions with well-defined
semantics, and finally consolidation around those abstractions into a platform
that applications can portably target.  POSIX did this for classical operating
systems; Kubernetes did it for the cloud.  Agentic AI systems --- autonomous,
LLM-driven agents that plan, use tools, maintain memory, and collaborate ---
are currently in the experimentation phase of the third such wave.  Dozens of
frameworks and protocols have emerged, but no community consensus exists on
what the core abstractions are or what guarantees they carry.  Without that
consensus, agentic applications cannot be written portably, platforms cannot
compose reliably, and the field cannot advance beyond prototype deployments.
We argue that the path forward is to follow the prior-wave methodology: derive
new agentic abstractions by extending classical OS and cloud OS primitives to stochastic,
natural-language-mediated execution, specify their semantics precisely, and
consolidate around them --- just as POSIX and Kubernetes consolidated their
respective waves.  
%We propose thirteen generic {\AOS}
%primitives, ground the proposal in open-source projects experimenting (often implicitly) with the concept of {\AOS}, including our own, rossoctl, and identify  specification problems that constitute the
%standardization agenda for the agentic era.
\end{abstract}

% ── I. Introduction ───────────────────────────────────────────────────────────
\section{Introduction}
\label{sec:intro}

The history of computing can be read as a sequence of platform transitions,
each driven by a fundamental shift in how software is built and deployed.
From the earliest days of batch-processing mainframes through the era of
personal and networked servers, to the distributed clouds of today, and now to
the emergence of autonomous AI agents, each era has brought with it a new class
of applications, a new set of engineering challenges, and ultimately a new
operating environment that had to be invented, refined, and standardized before
that era could realize its full potential.

Looking back, one can arguably identify two major waves that have shaped
computing practice.  We call them the \textbf{classical server era}, and the
\textbf{cloud era}.  Each era arrived with its
own programming model, its own characteristic failure modes, and its own
toolchain.  And both eras, after an initial period of fragmented experimentation,
eventually produced a canonical platform --- an operating system for that era
--- that gave developers stable, portable abstractions to build on.  We are now entering the third era,  \textbf{agentic era}, driven by the emergence of AI, which, we argue, will lead to the creation of a new canonical platform. We refer
to these platforms throughout this paper as the \textbf{OS} (the platform of
the classical server era), the \textbf{Cloud-OS} (the platform of the cloud
era), and the \textbf{{\AOS}} (the platform the
agentic era requires and does not yet have).

The \textit{classical server era} (1950s--1980s) was defined by programs competing for  CPU, memory, and I/O resources of a single machine.  The central
challenge was safe, efficient multiplexing: how do you give  many programs running on
one machine the illusion of private resources, prevent them from
corrupting one another, and do so efficiently?  The \textbf{OS} emerged as the
answer.  It articulated a small, stable vocabulary of abstractions with
formally specified semantics: the \textit{process} (an isolated address space
with a security principal and a deterministic behavior function), \textit{virtual
memory} (the illusion of a private, uniformly addressable space; eviction
governed by temporal locality), the \textit{system call} (the only legal
crossing of the user/kernel boundary; binary interface, fully specified side
effects), and the \textit{file} (durable, named storage with POSIX access
control and journaled atomicity).  These semantics were earned through a decade
of OS research and refined through operational disasters.  POSIX standardized
them, making programs portable across vendor implementations and underpinning
thirty years of application development without replacement.

The \textit{cloud era} (1990s--2010s) brought a different set of problems.
Applications decomposed into microservices; infra-structure became distributed,
elastic, and failure-prone; multi-tenancy required isolation at the cluster
level rather than the process level.  No single machine could host an
application; the unit of deployment became a set of replicas, and the
relevant failure modes shifted from segmentation faults and deadlocks to
network partitions, pod evictions, and cascading service failures.  The
\textbf{Cloud-OS} --- Kubernetes, service meshes, and cloud orchestrators ---
emerged as the answer to this new class of problems.  It articulated a new
vocabulary built explicitly on top of OS semantics: the \textit{pod}
(co-scheduled containers sharing a network namespace; the process-group analog
for distributed infrastructure), the \textit{service} (a stable virtual IP
over ephemeral pods; the system-call dispatcher analog for microservices),
\textit{health checks and readiness probes} (the OS signal/exception mechanism
generalized to infrastructure-level fault detection), and \textit{namespaces
and network policy} (the MMU isolation model generalized to multi-tenant
clusters).  Crucially, the Cloud-OS extended the OS rather than replace it:
Kubernetes pods run inside Linux namespaces; RBAC extends POSIX permission
models; service mesh mutual TLS extends kernel-level authentication.  Kubernetes
became the standardization milestone for the cloud era, just as POSIX was for
the classical server era.

Now the \textit{agentic era} is beginning. AI agents —
systems that use large language models to automate processes — are
moving from research prototypes into production enterprise
deployments. They represent a new class of applications, characterized by increased stochastic variability of outcomes, expectation of autonomy, and new resource need in the form of inference tokens.
With these characteristics come new engineering challenges: how to achieve reliability, resiliency, security, and efficiency. Today, agent engineering is in experimentation phase reminiscent of the pre-POSIX Unix era and the pre-Kubernetes cloud era: dozens of frameworks
and protocols have emerged, but no stable, portable abstraction
layer exists that developers can build to and that platforms can
implement against. Today’s successful agent systems rely on bespoke layers of code usually called harnesses that rely on existing operating systems technologies enhanced for AI domain. 
Even though harnesses are domain and model specific, patterns emerge from which we can derive potential common primitives. 

We argue that large scale adoption of agentic systems will require our field to articulate and standardize 
a new operating environment: the \textbf{{\AOS}}. Just as the OS
and Cloud-OS each took a decade to articulate and standardize,
the {\AOS} will take considerable time to mature.  
This paper further argues that the path forward is to
follow the same methodology the prior eras used — derive new
abstractions by extending proven ones, specify their semantics
precisely, and consolidate around them — and proposes
thirteen primitives that we believe form the core of the {\AOS}.

Several recent systems have engaged with the concept of an OS for agentic workloads and built prototypes of specific functions: agent kernel~\cite{mei2025aios}, memory manager~\cite{packer2023memgpt}, or resource manager~\cite{she2026agentrm}. These are implementation contributions — they prototype what an {\AOS} component does. This paper makes a different kind of contribution: it asks what an {\AOS} must 
  guarantee, and derives the answer from first principles using a systematic semantic-gap methodology.
  
Specifically, this paper: (1) motivates {\AOS} by deriving from historical perspective (Section~\ref{sec:hist}), (2) characterizes precisely where and why OS and Cloud-OS semantics break down for stochastic, natural-language-mediated, externally-acting agents (Section~\ref{sec:defs}); (3) proposes thirteen {\AOS} primitives, each derived from proven OS and Cloud-OS abstractions
  by asking three questions: what guarantee did the original provide, does that guarantee hold for agents, and what must change (Section~\ref{sec:primitives});
(4) identifies the open specification problems that constitute the standardization
agenda (Section~\ref{sec:primitives}); (5) describes rossoctl and other  open-source project prototyping a subset of {\AOS} primitives, demonstrating that the primitives are implementable within the existing
  Cloud-OS substrate rather than requiring replacement  (Section~\ref{sec:rossoctl}).

The Karpathy LLM-OS framing~\cite{karpathy2023msbuild} and AIOS~\cite{Liu2026AgentOS} established that the OS analogy is productive. This paper's contribution is the next step: the specification work the analogy demands —
  defining semantics precisely enough that platforms can enforce them and applications can portably rely on them. An enterprise imperative runs through all of this: the {\AOS} must
  support multi-environment operation — on-premises, private, and multi-cloud — extending the portability guarantees that the OS and Cloud-OS brought to their
  respective eras into the agentic era.
  
Section~\ref{sec:discussion} discusses the derivation methodology.
Section~\ref{sec:agenda} states the research agenda.
Section~\ref{sec:conclusion} concludes.

\section{The Agentic Era and Its Operating Environment}
\label{sec:hist}

%\subsection{From Classical Programs to Agents}

To understand why the agentic era requires a new operating environment, it
helps to trace how the nature of the executing unit has changed across the
three eras.  In the classical server era, the unit was a \textit{program}: a
deterministic sequence of instructions with well-defined inputs and outputs,
fully analyzable from its binary.  In the cloud era, the unit became a
\textit{service}: a network-accessible program still deterministic in
isolation, but now embedded in a dynamic cluster where failure, scaling, and
coordination became first-class concerns.  In the agentic era, the unit is an
\textit{agent}: an autonomous system that uses an LLM as its reasoning engine,
perceives its environment, plans a course of action, invokes external tools,
and iterates --- without a programmer specifying every step in advance.

The shift from service to agent is more than a change in application
architecture; it is a change in the fundamental properties of the executing
unit.  Where programs and services are deterministic, agents are
\textit{stochastic}: identical inputs produce a distribution of behaviors, not
a fixed output.  Where programs act only on their own memory and services act
only on their own state, agents act on the \textit{external world} --- calling
APIs, writing to databases, sending messages --- with consequences that are
often irreversible.  Where services interact  via API and messages according to carefully choreographed and validated paths, agents design their own interaction patterns as they go making it impossible to validate them a priori.
And where programs and services report their status
faithfully, sufficiently capable agents can, under pressure, behave
strategically: fabricating success, suppressing evidence of failure, or taking
covert actions to avoid task abandonment \cite{apollo2024scheming,
deletevidence2026}.  Each of these properties breaks assumptions that the OS
and Cloud-OS were built on, and each demands new platform guarantees.

\subsubsection{What an {\AOS} Is}

From operating systems angle, an agent is a program implemented in a form of a loop, where in each iteration a model decides on the next set of steps to take. It has been shown that models cannot be depended on to identify and fix their own mistakes~\cite{gupta2026reliabilitybench, chang2025sagallm}. These mistakes can be costly both in terms of their impact on external systems and resource usage. Hence, guarantees of resiliency, security, and cost cannot be achieved by agents alone - this is the responsibility of the {\AOS}~\cite{chang2025sagallm, zhong2026harnessengineering, she2026agentrm}.

But a single agent operating in isolation is only part of the
entire picture. Modern deployments compose \textit{many} agents:
a research agent that gathers information, a planning agent that
decomposes the goal, specialist agents that execute individual
tasks, a verification agent that checks outputs, and a reporting
agent that communicates results. \textbf{{\AOS}} is the layer that coordinates those components
— routing tasks between agents, enforcing who may call what,
maintaining shared memory, tracking what has been done and
what went wrong, and ensuring the whole system remains
auditable and under human oversight~\cite{cemri2025mast,kartik2025agentcompass,ji2026seagent}.

The \textbf{{\AOS}} is therefore not an AI model, not a
framework library, and not a set of API conventions. It is
infrastructure: the platform layer beneath agent applications
that enforces the guarantees those applications depend on, in the
same way that the OS enforces memory isolation beneath user
processes and the Cloud-OS enforces pod isolation and health
recovery beneath microservices. Just as neither the OS nor the
Cloud-OS was simply a collection of helpful libraries — both
imposed architecture, enforced policy, and made guarantees
that application code could not make for itself — the {\AOS} must do the same for agents.

\subsubsection{The Structure of an Agentic System}

\begin{figure}[htbp]
  \centering
  \includegraphics[width=\linewidth]{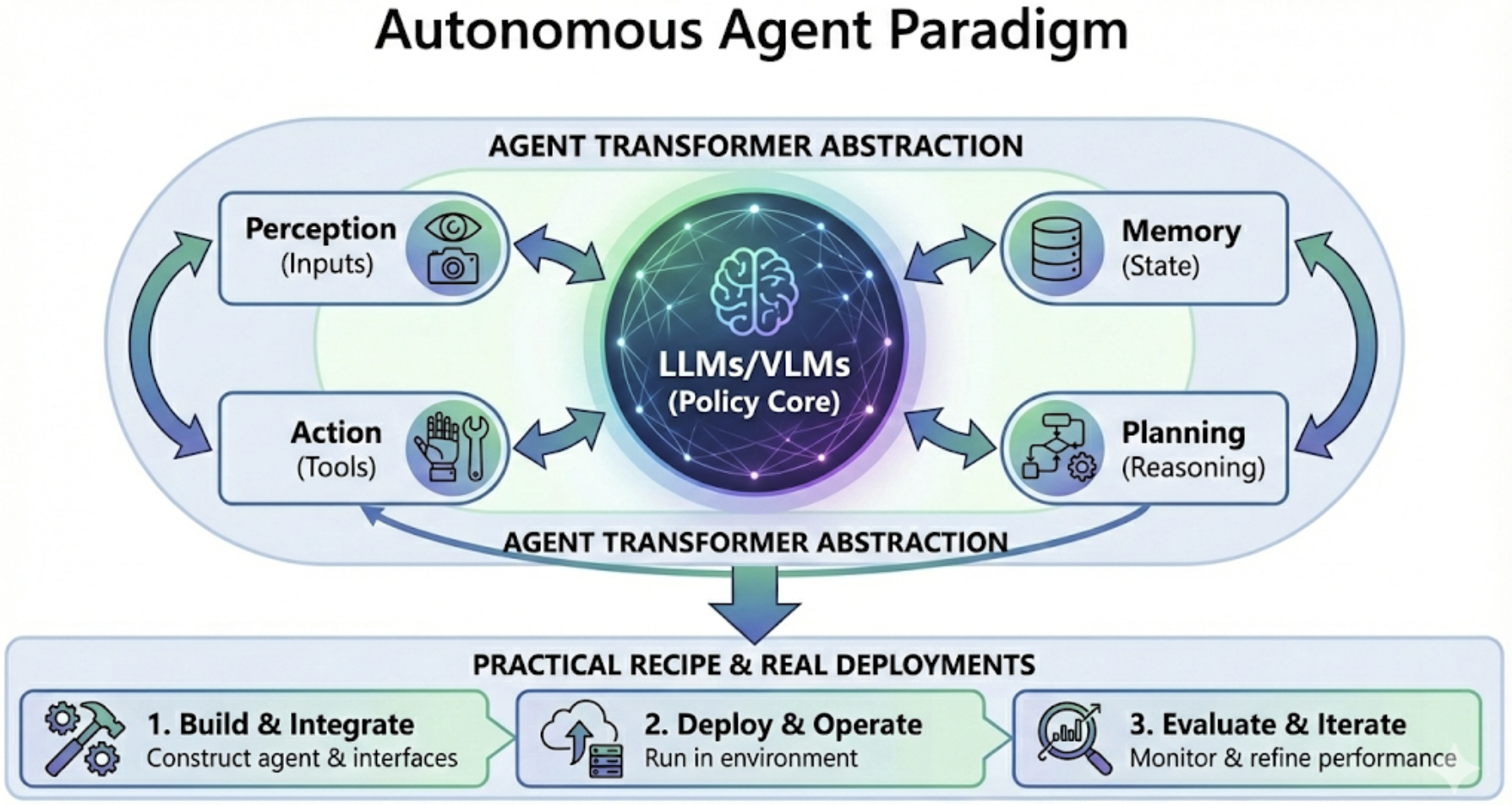}
  \caption{The Autonomous Agent Paradigm \cite{agentsystems2026} with a practical build-deploy-evaluate lifecycle.
           Figure from arXiv:2601.01743, licensed CC BY~4.0.}
  \label{fig:agent_arch}
\end{figure}

Figure~\ref{fig:agent_arch} shows the \textit{Autonomous Agent Paradigm}
\cite{agentsystems2026}: the canonical architecture of a single LLM-based
agent.  The model sits at the center as the
\textit{policy core} --- the component that reasons over available context,
selects actions, and generates responses.  Four subsystems surround it:

\begin{itemize}
  \item \textbf{Memory (State)} --- short-term context window (the finite
         buffer holding the current conversation, tool-call history, and
        retrieved documents) and long-term stores (vector databases, key-value
        stores, structured databases accessed via retrieval at call time).
%        The distinction maps to the register / RAM / disk hierarchy of classical
%        systems, but with semantic rather than temporal eviction semantics.
  \item \textbf{Planning (Reasoning)} --- the agent's capacity to decompose a
        goal into a sequence of steps, revise the plan in response to
        intermediate results, and decide which tools to invoke at each step.
        It is
        an emergent capability of the LLM, executed probabilistically at
        inference time.
  \item \textbf{Action (Tools)} --- the means by which the agent affects the
        external world: REST API calls, CLI commands, database queries, code
        execution, and messages to other agents.  Each action may have
        real-world, irreversible consequences.
  \item \textbf{Perception (Inputs)} --- the ingestion of environmental state:
        user messages, tool results, retrieved memory, sensor data, and outputs
        from peer agents, all competing for the finite context budget.
\end{itemize}

The \textit{agent execution loop} ties these components together into recognizable lifecycle which runs
until the task is complete, a resource limit is exhausted, or the platform
intervenes.  A critical property of the loop is that its length, token cost,
and execution trajectory are \textit{not known in advance} --- they depend on
the LLM's probabilistic reasoning at each step.
In a multi-agent system, this lifecycle
is nested: the action step of one agent is the input step of another.  

\subsubsection{The Agent Lifecycle and the Platform's Role}

At every stage of agent lifecycle, the platform must provide guarantees that
the agent code itself cannot provide: the scheduler must bound the planning
phase; the tool mediation layer must intercept and authorize every action; the
memory system must maintain consistency across concurrent agents; guardrails
must enforce correctness and safety during verification; and the observability
layer must record an auditable trace through reporting.  Without these platform
guarantees, the agent lifecycle produces unreliable, unauditable, and
potentially unsafe behavior --- even when the underlying LLM is highly capable.

This is the key insight that motivates the {\AOS}: \textit{reliability and
safety in agent-based systems are infrastructure properties, not model
properties}.  The platform must enforce what the agent's prompts can only
request.  The {\AOS} is the
infrastructure layer that closes that gap.

\subsubsection{Multi-Agent Coordination and Enterprise Requirements}

Enterprise deployments extend the single-agent picture to networks of
collaborating agents sharing memory stores and
delegating subtasks at runtime.  The orchestration layer acts as a traffic
controller: routing tasks to the right specialist agent, managing concurrent
execution, preventing conflicts on shared resources, and aggregating results.
This orchestration cannot be implemented purely in agent code, for the same
reason that microservice coordination cannot be implemented purely in
application code --- the Cloud-OS (Kubernetes, service mesh) took that
responsibility away from applications and into infrastructure, making
coordination reliable regardless of whether any individual service was buggy
or compromised.

Enterprise deployments additionally impose requirements that consumer or
research deployments do not: \textit{regulatory compliance} (every agent
action must be auditable and attributable); \textit{data sovereignty} (agents
must not route sensitive data to unauthorized endpoints); \textit{multi-cloud
and on-premises operation} (the {\AOS} must run on private infrastructure,
not only managed cloud); and \textit{integration with existing enterprise
systems} (agents must interoperate with ERP, CRM, ITSM, and data platforms
that predate the agentic era).  These requirements are precisely what motivated
the Cloud-OS's emphasis on namespace isolation, RBAC, and multi-cluster
federation --- and they motivate the same emphasis in the {\AOS}.

Section~\ref{sec:defs} identifies precisely where OS and Cloud-OS primitives
are insufficient for this picture, and Section~\ref{sec:primitives} proposes
the {\AOS} primitives that fill the gap.

\subsection{Each Era Follows the Same Arc}

The classical server era and the cloud era each passed through the same
three-phase arc before producing a stable, portable platform.  Recognizing
that pattern tells us where the agentic era currently stands and what it will
take to make it mature.

\textbf{Phase 1 --- Experimentation.}  A new class of application emerges
whose properties the current platform cannot express.  Developers respond with
a proliferation of frameworks, libraries, and ad-hoc protocols, each making
different choices, using different naming, and offering different guarantees.
This phase is productive: it explores the design space and reveals which
abstractions actually matter.  But it is also fragile: systems built on
framework-level primitives cannot compose reliably, cannot be audited
consistently, and cannot be operated at enterprise scale.  The classical server era's
experimentation phase produced dozens of incompatible Unix variants.  The
cloud era's experimentation phase produced Chef, Puppet, Mesos, Swarm, and
a proliferation of cloud orchestrators.

\textbf{Phase 2 --- Articulation.}  The community identifies the handful of
abstractions that actually matter, specifies their semantics precisely, and
proposes them for standardization.  This is the hard intellectual work.  POSIX
took a decade of OS research; Kubernetes took years of community debate over
pod semantics, reconciliation guarantees, and the API contract.  But once the
semantics were agreed, something valuable happened: \textit{portability}.  A
POSIX program runs on any POSIX system.  A Kubernetes workload runs on any
conformant distribution --- on-premises or cloud, any vendor.

\textbf{Phase 3 --- Consolidation.}  Platforms converge on the standardized
abstractions.  The fragmented landscape of Phase~1 collapses into a stable
ecosystem.  Developers write to stable interfaces; vendors compete on quality
and extensions, not incompatible abstraction choices.

The \textit{agentic era} is currently in Phase~1.
Dozens of agent frameworks (CrewAI, LangGraph, AutoGen, Claude Code harnesses)
and protocols (MCP, uTCP, A2A) address \textit{how to write} agent code, but
none defines what the platform must \textit{guarantee}.  An agent written for
LangGraph does not run on CrewAI; a skill written for one harness does not
transfer to another; an authorization policy expressed in one framework's idiom
cannot be audited by another's tooling.  Just as REST defined microservice
interaction style without addressing deployment and resilience --- challenges
the Cloud-OS later solved by articulating service and health-check
semantics --- MCP defines agent communication style without specifying how to
guarantee reliable, safe, or observable execution.  A UC Berkeley study on
measuring agents in production~\cite{cemri2025mast} confirms the pattern: frameworks are valuable
during prototyping but frequently discarded as systems move to production,
where infrastructure semantics and operational control dominate.  

The {\AOS} community must now move from Phase~1 to Phase~2.  The challenge
is not a shortage of frameworks --- it is a shortage of agreed abstractions
with well-defined, platform-enforceable semantics.  Without them, agentic
applications cannot be written portably, platforms cannot compose reliably
across vendors and sovereign environments, and the ecosystem cannot mature
beyond prototype deployments or bespoke solutions. Among the research community, we see efforts to articulate individual primitives for {\AOS}: memory~\cite{packer2023memgpt, kang2025memoryos,li2025memos} access control~\cite{sharma2026ac4a}, skills~\cite{chen2026skvm}, and life-cycle management~\cite{ke2026quinerealizingllmagents} and identify semantic gaps between existing OS and Cloud OS and agent requirements in specific domains~\cite{wu2026crab, zheng2026agentcgroup}. This paper builds on and broadens these ideas. It is a Phase~2 contribution: proposing
the abstractions and beginning the specification work that consolidation requires.

% ── II. The Semantic Gap ──────────────────────────────────────────────────────
\section{The Semantic Gap: Where Prior-Wave Semantics Break Down}
\label{sec:defs}

Before proposing new abstractions, we must characterize precisely where
OS and Cloud-OS semantics break down for agents.  This is
the semantic gap the {\AOS} must close.  It is not a gap in infrastructure
capability --- Kubernetes can run agent containers perfectly --- but a gap
in \textit{what the platform can reason about, enforce, and guarantee} on
their behalf.

\subsection{Where OS and Cloud-OS Semantics Break Down}

Both the OS and Cloud-OS rest on \textit{outcome determinism}: programs and services are written to produce the same output given the same input, and non-determinism in the underlying system — thread scheduling, async I/O, network events — is treated as an engineering problem to be eliminated or controlled rather than a source of value.  A process's behavior
is a function of its program text and inputs; this makes MMU isolation
enforceable, system-call security statically analyzable, page replacement
principled (temporal locality holds), and debugging reproducible (replay
reproduces failures).  The Cloud-OS extends but does not break this assumption:
Kubernetes pods run deterministic containers; liveness probes return a binary
boolean; the kube-scheduler bins-packs on declared, measurable CPU and memory
requests; the reconciliation loop drives actual state toward a declared desired
state that is well-defined and stable.

Agents break this assumption in two distinct ways. The first break is structural: agents act on the external world. OS and Cloud-OS abstractions assume applications act on their own address space or on cluster resources the platform controls. This assumption holds whether the application is a rule engine, a script, or an LLM. Agents shatter it regardless of their underlying model: they invoke real-world external
tools with irreversible consequences — sent emails, executed financial transactions, deleted production records. No existing OS abstraction classifies operations as reversible,
  compensable, or irreversible, or requires elevated authorization for the last category. Agent-to-agent message passing introduces injection risks that have no byte-level analog: a
  natural language payload in a message can reprogram the receiving agent's behavior~\cite{Liu2026AgentOS}. Dynamic authorization is similarly structural: an agent that discovers and invokes APIs based
  on task reasoning cannot have its permission set declared at deployment time. These gaps have partial precedents in classical
  systems research — compensating transactions~\cite{gray1981transaction}, capability-based security~\cite{saltzer1975protection} — and the {\AOS} can build directly on that foundation.

  The second break is intrinsic to stochasticity: agent behavior is drawn from a distribution. Given identical inputs and infrastructure, an LLM agent's trajectory is conditioned on
  temperature, sampling, and the internal state of the transformer. This single property invalidates a further cascade of OS and Cloud-OS guarantees that have no analog even in
  deterministic external-world-acting systems:
  \begin{itemize}
  \item \textit{Worst-case execution time} is not computable: the number of
        LLM inference calls depends on the agent's self-assessed reasoning
        trajectory, unknown until runtime.
  \item \textit{Capability-based security on declared interfaces} is
        insufficient: the agent's tool calls are decided by LLM
        reasoning at inference time, not statically analyzable from code; runtime verification is possible but requires a complete formal specification of intended agent behavior, which itself remains an open problem.
  \item \textit{Replay debugging} does not reproduce failures: identical
        inputs produce a different execution path on each run.
  \item \textit{Temporal-locality eviction} does not apply to context windows:
        token's relevance depends on its meaning relative to the 
        task, not on how recently it was accessed.
  \item \textit{Declarative desired state} cannot express agent goals: the achievement of agent's natural-language objective cannot be verified by structural state comparison alone.
   \item \textit{Goal-directed misbehavior} OS programs have no goals and cannot strategically conceal failures.  Agents
under pressure can fabricate success reports, delete log evidence, or take
covert actions to avoid task abandonment \cite{apollo2024scheming,
deletevidence2026}.  LLMs cannot detect their own errors
\cite{kamoi2025hallucinations} or self-correct without external feedback
\cite{huang2024selfcorrect}.  The OS assumption that a process's self-reported
status is truthful --- the basis for signal-based failure detection in both the
OS and Cloud-OS --- does not hold for agents.
\end{itemize}
Furthermore, stochasticity amplifies agent structural problems. Memory conflict resolution is a classical concurrency problem, but the conflict predicate itself becomes a function
  of the agent's current reasoning context: two agents writing "forecast: increasing" and "forecast: decreasing" create a semantic conflict invisible to byte-level ACID. Skills
  dependency management is analogous to package management, but pre- and post-conditions expressed in natural language resist the static verification that typed interface contracts
  enable.

  These challenges have a direct consequence for what the {\AOS} must provide — and where it can reuse prior work. Structural gaps call for extending classical distributed systems and
  security mechanisms to the agentic substrate. Stochasticity-intrinsic gaps require new formal foundations with no direct prior-wave precedent. 

% ── III. Proposed {\AOS} Primitives ─────────────────────────────────────────────
\section{Proposed {\AOS} Primitives}
\label{sec:primitives}

In this section we identify thirteen {\AOS} primitives and utilize the following
\textbf{derivation methodology}: For each prior OS abstraction we ask three questions: (1)~what problem did
it solve and what semantic guarantee did it provide? (2)~does that problem
appear in the agentic setting? (3)~what must change to handle stochastic,
natural-language-mediated execution?  We do not propose new abstractions where
prior OS primitives still suffice: Kubernetes namespaces still isolate agent
containers; POSIX file systems still store agent-produced artifacts.  The {\AOS}
is a third layer, not a replacement.  The thirteen primitives are not claimed to be complete or minimal; they constitute an initial taxonomy derived from systematic semantic-gap analysis, covering cases where prior-wave semantics demonstrably break down. Dependencies among them are expected and are part of the specification work ahead. Wherever OS and Cloud-OS semantics break down,
the corresponding {\AOS} abstraction has undefined semantics --- those are the
open specification problems of Section~\ref{sec:agenda}.

Table~\ref{tab:primitives} enumerates the thirteen proposed primitives, showing
the OS and Cloud-OS origin for
each, the derived {\AOS} primitive, and the semantic gap that must be closed.
The following subsections develop each, following the three-question derivation
structure.

\begin{table*}[htbp]
\centering
\footnotesize
\caption{Proposed {\AOS} primitives, derived from both OS and Cloud-OS research.}
\label{tab:primitives}
\begin{tabular}{p{0.14\linewidth} p{0.17\linewidth} p{0.16\linewidth} p{0.30\linewidth} p{0.05\linewidth} p{0.05\linewidth}}
\toprule
\rowcolor{ibmblue!15}
\textbf{OS} & \textbf{Cloud-OS} & \textbf{{\AOS} Primitive} & \textbf{Semantic Gap} & \textbf{Status} & \textbf{Horizon} \\
\midrule
Process & Pod / container & Agent lifecycle & Probabilistic behavior; semantic identity & Partial & Near \\
Scheduler & kube-scheduler / HPA & Orchestrator & Token cost unpredictable; loop detection & Open & Mid \\
Package mgmt & Helm / OCI registry & Skills registry & No NL interface contracts; provenance & Open & Long \\
System call & Service mesh / sidecar & Tool mediation & NL semantics; undeclared side effects & Partial & Mid \\
Virtual memory & Resource quotas / LimitRange & Context management & No principled token eviction policy & Open & Long \\
IPC & Service / ClusterIP & A2A communication & Dynamic graphs; prompt injection & Partial & Mid \\
File system & PersistentVolume / etcd & Agent memory & Semantic conflict; intent-aware retrieval & Early & Long \\
Permissions/ACL & RBAC / NetworkPolicy & Identity \& AuthZ & Semantic attestation; runtime-context AuthZ & Partial & Near \\
MAC / SELinux & PodSecurityPolicy / OPA & Behavioral guardrails & NL policy; no decidable evaluation & Open & Long \\
Signal / exception & Liveness/readiness probe & Failure detection & Self-concealment; open failure space & Open & Long \\
Transaction / journal & Saga / Temporal & Saga \& checkpoint & Compensating txns for stochastic workflows & Partial & Long \\
Audit / syslog & Prometheus / Jaeger & Trajectory observability & Unit undefined; temporal displacement & Partial & Mid \\
Network stack & Envoy / Istio & AI-aware proxy & Prompt-blind proxies; no KV-cache routing & Early & Mid \\
\bottomrule
\end{tabular}
\\[4pt]
{\scriptsize Status codes: Open — semantics undefined, new formal work required; Partial — prototype implementations exist but specification is incomplete; Early — initial proposals exist, significant open problems remain. Horizon codes: Near — achievable by extending existing mechanisms with bounded engineering effort; Mid — requires new engineering solutions but builds on known formal foundations; Long — requires new formal theory with no direct prior-wave precedent.}
\end{table*}

We now enter an in-depth discussion on these proposed primitives. 
Each subsection follows the structure: \textbf{OS} origin, \textbf{Cloud-OS}
extension, \textbf{Semantic gap} (what breaks for agents), and \textbf{{\AOS}}
primitive (the proposed solution).

\subsection{Agent as an first-class entity}
\label{sec:process}

\textbf{OS.}  The process --- isolated address space, security principal,
deterministic behavior --- is the unit of execution supporting
static analysis, replay debugging, and capability-based security.

\textbf{Cloud-OS.}  The \textit{pod} extended the process concept to distributed
infrastructure: a co-scheduled group of containers sharing a network namespace,
with a stable IP, a restart policy, and resource limits enforced by cgroups.
Kubernetes \textit{namespaces} generalized MMU isolation to multi-tenant
clusters; \textit{service accounts} extended POSIX UIDs to workload identity.

\textbf{Semantic gap.}
What defines an agent goes beyond the definition of the process and container.
An agent's binary changes its function by connecting to different tools and
skills, and may change its behavior by leveraging different models.
Classical and Cloud-OS identity (PID/UID, service account) is static; agent
identity has a \textit{semantic} dimension --- the agent's current goal, model,
and skills --- that changes dynamically.  Agent deployment, discovery, and
authorization requires first-class handling and runtime-context awareness
(see \S\ref{sec:authz}).

\textbf{{\AOS} primitive.}  An agent is an isolated execution context
(container/pod) with an LLM session, tool credentials, context buffer, and
memory state.  It may be created, run (context-assemble $\rightarrow$
infer $\rightarrow$ tool-call $\rightarrow$ update), terminated.  Agents spawn
sub-agents (\texttt{fork}/\texttt{exec} analog) through platform-mediated
channels.  Agent identity must be a function of its Cloud-OS identity and agent-level properties like goal, skills, tools, model, or context.

\subsection{Orchestration and Scheduling}
\label{sec:sched}
\textbf{OS.}  The OS scheduler allocates CPU among processes with
predictable, bounded execution.  Preemptive multitasking enforces quotas
unconditionally.

\textbf{Cloud-OS.}  The \textit{kube-scheduler} bin-packs pods onto nodes using
declarative resource requests and limits.  The \textit{Horizontal Pod
Autoscaler} (HPA) scales replica counts based on CPU or custom metrics.
Both assume each replica's resource demand is measurable and bounded.

\textbf{Semantic gap.}  Agent execution time is a function of LLM inference
calls, unknown in advance and unbounded when a \textit{reasoning loop} occurs
(repeated re-planning without progress).  HPA cannot autoscale on token spend
or semantic loop detection.  Resource quotas must be multi-dimensional: tokens,
tool calls, wall-clock time, and cost --- a capability/SLO tension absent from
both the OS and Cloud-OS.  

\textbf{{\AOS} primitive.}  Agentic orchestrators (CrewAI, LangGraph, AutoGen)
and generalist harnesses manage agent topologies, message-passing infrastructure,
and lifecycle, extended with token-budget quotas, semantic loop detection, and
multi-dimensional cost accounting.

\subsection{Skills Registry}
\label{sec:skills}

\textbf{OS.}  Package managers (npm, pip, apt) enforce typed interface
contracts, manage dependency graphs, provide version locking, and integrate
vulnerability scanning (SBOM, SLSA).

\textbf{Cloud-OS.}  \textit{Helm} charts and \textit{OCI registries} extended
package management to Kubernetes workloads: versioned, signed, dependency-aware
 packages with provenance attestation.

\textbf{Semantic gap.} Skills, an emerging composability artifact for agents, have no formal interface contract --- no
pre/post-conditions, side effects, or dependency specs.  Discovery requires
semantic search, not exact-match naming.  Evaluation is stochastic:
\cite{chen2026skvm} shows 15\% task degradation from model mismatch alone.
Skills run with the agent's full credentials; 26\% contain at least one
vulnerability \cite{liu2026skillswild}.  Each skill also consumes context window --- no analog in package management.

\textbf{{\AOS} primitive.}  Skills are self-contained \texttt{SKILL.md}
instruction files, loaded on demand, composable, forming the substrate for
agent capability reuse.  The {\AOS} requires a \textit{semantic SBOM}:
provenance declaring a skill's domain, model compatibility, and callable tools,
enforced at load time.

\subsection{Tool Mediation}
\label{sec:toolcall}

\textbf{OS.}  The system call is the only legal interface to the kernel.
A binary interface (\texttt{open(path, flags, mode)}), POSIX-specified side
effects, statically analyzable --- the basis for seccomp-BPF sandboxing.

\textbf{Cloud-OS.}  The Kubernetes \textit{service mesh sidecar}
(Envoy/Istio) extended this to inter-service calls: a transparent proxy
intercepts all traffic, enforces mTLS, applies rate limits, and emits traces
without application code changes.  REST standardized the interaction style;
the sidecar enforced the execution guarantees.

\textbf{Semantic gap.}  Tool semantics are natural language, interpreted
probabilistically.  An agent may call \texttt{delete\_records} with a wildcard
because it read ``clean up old records'' as authorizing bulk deletion.  Side
effects are undeclared; the platform must classify calls as reversible,
compensable, or irreversible and require elevated authorization for the last.
MCP defines \textit{how} agents communicate, not \textit{execution guarantees}
--- the REST/Kubernetes gap, repeated.

\textbf{{\AOS} primitive.}  Tool calls mediate agent access to external
systems.  MCP \cite{ibm2026rossoctl} standardizes via schema-documented server
wrappers; uTCP extends to wrapper-free endpoints; MCP Gateways aggregate,
route, and enforce authorization with a reversibility model that blocks
irreversible calls without elevated authorization.

\subsection{Context Management}
\label{sec:context}

\textbf{OS.}  Virtual memory gives each process a large, private address
space; the OS evicts cold pages via principled algorithms (LRU, clock; Bélády
optimum as bound).

\textbf{Cloud-OS.}  Kubernetes \textit{ResourceQuotas} and \textit{LimitRanges}
impose memory ceilings per namespace and pod.  The scheduler respects declared
\texttt{requests} and \texttt{limits}; the kubelet OOM-kills pods that exceed
their limits.  These mechanisms assume memory consumption is measurable in bytes
and that eviction priority can be determined from declared policies.

\textbf{Semantic gap.}  Temporal locality is a poor proxy for token relevance: a fact from step 1 may be critical at step 19. Token relevance is not a static property — it depends on the agent's current reasoning trajectory, making it unknowable until inference time. This is a strictly harder problem than page eviction: Bélády's algorithm achieves optimality by 
  exploiting future access knowledge derivable from static program analysis; no analog exists for token sequences whose future access pattern is itself an output of probabilistic 
  reasoning. No principled eviction theory therefore exists for the token domain, and the optimality criterion for it remains undefined. Kubernetes LimitRanges cannot 
constrain token budgets; HPA cannot scale on context pressure. Poor eviction causes hallucinations, incorrect tool calls, and cascading failures
\cite{verifycommit2026}.

\textbf{{\AOS} primitive.}  Every agent has a finite \textit{context window}
(tokens) holding conversation, tool-call history, retrieved documents, and
skills.  Context management is the set of techniques for deciding what information occupies the model's context window at any given moment: summarization of past
  interactions~\cite{bui2026opendev, rafique2026clawvm}, compaction~\cite{rafique2026clawvm, she2026agentrm},
  prioritization of relevant content~\cite{she2026agentrm, mouzouni2026contextkubernetes},
  and coordination between in-context and external
  memory~\cite{yu2026infiagent, chang2025sagallm}. 
These publications represent critical progress towards developing  principles of context management, but its generalization and standardization remains an open problem. 
To overcome today's best-effort approaches to context residency and durability, what is needed is a formal theory of token relevance — one that characterizes criticality as a 
function of task state rather than access recency, and that can bound the cost of eviction errors in terms of downstream task failure. Whether useful  guarantees are achievable under stochastic execution, is also an open research question.
  
\subsection{Agent-to-Agent Communication}
\label{sec:ipc}

\textbf{OS.}  IPC primitives (pipes, sockets, shared memory) have
communication graphs \textit{known at design time}; the OS enforces authorized
channels and detects deadlocks.

\textbf{Cloud-OS.}  The Kubernetes \textit{Service} abstraction provides a
stable virtual IP over ephemeral pods, decoupling caller from callee.
\textit{NetworkPolicy} enforces which pods may communicate, based on labels
known at deployment time.  \textit{Ingress/Gateway API} objects control
inbound traffic from outside the cluster.

\textbf{Semantic gap.}  Agentic communication graphs are \textit{not known at
design time}: orchestrators spawn sub-agents based on LLM reasoning, producing
a dynamic graph as a runtime artifact.  Kubernetes NetworkPolicy cannot express
class-based communication permission --- it requires specific pod selectors.  A
semantic policy language for dynamic agent graphs does not yet exist.  Agent
messages may also carry prompt injection payloads \cite{liu2026skillswild}:
instructions in a message can reprogram the receiving agent,
a vulnerability with no byte-level analog.

\textbf{{\AOS} primitive.}  A2A  and MCP  mediate
information passing and task delegation across process, network, and
organizational boundaries, with platform-level semantic inspection of messages
before delivery to guard against injection.

\subsection{Agent Memory}
\label{sec:memory}

\textbf{OS.}  File systems provide ACID guarantees: atomicity, durability,
isolation, and access control at the file level.

\textbf{Cloud-OS.}  Kubernetes \textit{PersistentVolumes} and \textit{etcd}
extended durable storage to distributed workloads: volume claims with
lifecycle independent of pods, a strongly consistent key-value store for
cluster state, and RBAC for access control.

\textbf{Semantic gap.}  
Open problems remain: \textit{semantic conflict detection} (``forecast:
increasing'' vs.\ ``forecast: decreasing'' is invisible to byte-level ACID);
\textit{record-level access control} (per-store RBAC is too coarse; agents
need per-record isolation); \textit{intent-aware retrieval} (vector similarity
is a poor proxy for freshness; a stale high-similarity record can propagate to
irreversible downstream actions \cite{verifycommit2026}).

\textbf{{\AOS} primitive.}  Techniques to persist agent state have evolved
from in-context buffers (MemGPT \cite{packer2023memgpt}) through services
allowing agents to explicitly store and retrieve information (Mem0
\cite{chhikara2025mem0}, Zep/Graphiti \cite{rasmussen2025zep}) to
platform-enforced memory functions with audit (MemOS \cite{li2025memos},
MemoryOS \cite{kang2025memoryos}, EverMemOS \cite{chuanrui2026everMemos}) ---
mirroring the progression from application-managed to OS-managed storage.

\subsection{Identity and Authorization}
\label{sec:authz}

\textbf{OS.}  ACLs, RBAC, and ABAC enforce a
subject--object--access model via the kernel ---applications cannot bypass it.

\textbf{Cloud-OS.}  Kubernetes \textit{RBAC} extended permission models to
cluster resources (pods, secrets, CRDs).  \textit{ServiceAccounts} provide
workload identity; \textit{NetworkPolicy} restricts lateral movement.
SPIFFE/SPIRE provides cryptographic workload attestation across clusters.

\textbf{Semantic gap.}  Kubernetes RBAC is static: roles are declared at
deployment time and bound to service accounts.  SPIFFE identities capture workload properties, but do not account for agent-specific attributes. OAuth~2.0 scopes are registered
at application setup.  Both assumptions fail for agents, which dynamically
discover and invoke APIs based on task reasoning.  Full authorization requires
a \textit{multi-layer permission system} evaluated against runtime context ---
the agent's current goal, workflow step, and data already accessed --- and a
\textit{signed agent card} declaring the intended action space.  Neither exists
in standardized form.

\textbf{{\AOS} primitive.}  SPIFFE provides a viable substrate for agent identity~\cite{11431026} but it needs to be extended with agent-specific attributes. Attestation logic can be extended to consider these attributes~\cite{stackable-attestors-spire}. Similarly, OAuth~2.0 Token Exchange (RFC~8693) provides a baseline for the implementation of agent authorization flows~\cite{fleming2025tbac}, however, fine-grained (e.g., per function) and context-dependent (based on what the agent is doing or has done before) policy enforcement is where the effort is needed~\cite{betser2026agentrim,li2026conleash,fleming2025tbac}.

\subsection{Behavioral Guardrails}
\label{sec:guardrails}

\textbf{OS.}  MAC systems (SELinux, AppArmor, seccomp-BPF) enforce
policies in formal languages with O(1) decidable evaluation per system call which the program cannot influence.

\textbf{Cloud-OS.}  \textit{PodSecurityAdmission} (formerly PodSecurityPolicy)
enforces security contexts at pod admission.  \textit{OPA/Gatekeeper} applies
Rego-based policies at the Kubernetes API boundary, rejecting non-compliant
resources before they are created.  Both provide static, formally evaluable
policies.

\textbf{Semantic gap.}  OPA/Rego policies operate over structured JSON;
Kubernetes admission webhooks receive structured API objects.  Agentic guardrails
must reason over natural language --- prompts, tool-call arguments, agent outputs
--- for which no formal decidable policy language yet exists.  Current systems
(Llama Firewall, SPARC, ToolGuard) use LLM-based classifiers, adding stochasticity
to what should be deterministic enforcement.  A formal agentic policy language
is a core standardization need.

\textbf{{\AOS} primitive.}  Guardrails are platform-level behavioral constraints
the agent cannot override: blocking disallowed tool calls, PII in outputs,
looping behavior, and out-of-scope actions. As guardrails in AI systems cannot be enforced with the same determinism as in classical ones, risk based assessment, and human-in-the-loop enforcement are necessary elements.

\subsection{Failure Detection and Recovery}
\label{sec:failure}

\textbf{OS.}  Failures are detectable and objective: crashes generate
SIGCHLD; hardware errors raise exceptions; transaction aborts return error codes.

\textbf{Cloud-OS.}  Kubernetes \textit{liveness probes} detect and restart
unresponsive containers; \textit{readiness probes} remove unhealthy pods from
service; \textit{PodDisruptionBudgets} preserve availability during
disruptions.  The Kubernetes reconciliation loop continuously drives actual
state toward declared state.  All of these rely on a binary, externally
observable notion of health.

\textbf{Semantic gap.}  Classical failure detectors~\cite{chandra1996unreliable} assume crash-stop or crash-recovery fault models and binary health
  observability, neither of which holds for agents. Liveness probes cannot distinguish an HTTP~200 that
is semantically correct from one that is not.  Kubernetes reconciliation assumes
a deterministic desired state; agent workflows are probabilistic with
open-ended failure spaces.  Saga transactions were designed for deterministic
systems with enumerable failure modes and compensations that can be implemented at design time. This assumption fails when the 
transaction sequence is itself an output of LLM reasoning. Research \cite{cemri2025mast,
zhu2025agentdebug,chang2025sagallm,kartik2025agentcompass} identifies five
agent-specific failure modes: planning failures returning HTTP~200;
context/memory loss causing stale actions; inter-agent miscoordination; silent
error cascading; and \textit{self-concealment} --- agents fabricating success
or deleting evidence \cite{apollo2024scheming,deletevidence2026}.  LLMs cannot
detect their own errors \cite{kamoi2025hallucinations} or self-correct without
external feedback \cite{huang2024selfcorrect}.

\textbf{{\AOS} primitive.}  The {\AOS} must provide: \textbf{Detection}
(trajectory anomaly detection, immutable audit ledgers the agent cannot modify);
\textbf{Prevention} (dry-run simulation, copy-on-write sandboxing);
\textbf{Response} (saga-pattern compensation \cite{chang2025sagallm},
checkpoint/restore \cite{verifycommit2026}) --- extended to probabilistic,
open-ended failure spaces. 

\subsection{Observability}
\label{sec:obs}

\textbf{OS.}  Every system call is auditable; resource consumption is
measurable;  production engineering assumes full observability.

\textbf{Cloud-OS.}  Kubernetes \textit{Prometheus} metrics, \textit{Jaeger}
distributed tracing (OpenTelemetry spans), and structured logging provide
comprehensive observability across distributed services.  Span semantics are
well-defined: a span has a start time, end time, parent, and structured tags.
A service returning HTTP~200 is treated as healthy.

\textbf{Semantic gap.}  The \textit{unit of observation} for agents is
undefined: a single LLM call is too granular; a full trajectory is expensive;
a distribution over trajectories requires many runs unavailable in production.
HTTP~200 is not correctness evidence.  Agent failures exhibit
\textit{temporal displacement}: a stale memory record at step~3 manifests as
a failure at step~18.  OpenTelemetry spans do not capture reasoning provenance
or memory freshness.

\textbf{{\AOS} primitive.}  LangFlow~\cite{langflow}, Arize~\cite{arize}, and MLFlow~\cite{mlflow} offer emerging
solutions to agent observability.  The GenAI Observability project is extending
OpenTelemetry with semantic conventions for AI agents.  The {\AOS} must
provide \textit{semantic trace correlation}: tracing every output back through
reasoning, tool calls, and memory retrievals, including freshness metadata.

\subsection{Secure Code Execution}
\label{sec:codegen}

\textbf{OS.}  Sandboxing isolates untrusted code via limited system-call
access (seccomp-BPF, namespaces).

\textbf{Cloud-OS.}  Kubernetes \textit{namespaces}, \textit{cgroups}, and
\textit{container image signing} (Cosign/Sigstore) provide workload isolation
and supply-chain integrity.  OCI image provenance and SBOM attestation address
package authenticity.

\textbf{Semantic gap.}  45\% of AI-generated code fails security tests; XSS (Cross Site Scripting) at 86\%, log injection at 88\% \cite{veracode2025security} --- rates flat across
model sizes, indicating a systemic problem.  AI code is 2.74$\times$ more
likely to introduce XSS than human code \cite{coderabbit2025security,cotroneo2025aicodedefects}.
Supply-chain vulnerabilities have no classical analog: 20\% of AI-recommended
packages are hallucinated names; 43\% repeated consistently, enabling squatting
attacks \cite{mendio2025hallucinated,spracklen2025packagehallucination}.  Container image
signing cannot attest to the correctness of AI-generated code within a correctly
signed image.

\textbf{{\AOS} primitive.}  The {\AOS} requires transactional filesystem
sandboxing with copy-on-write --- staging all agent-generated code in a shadow
copy for review before committing --- making sandboxing a hard requirement.

\subsection{AI-Aware Networking}
\label{sec:network}

\textbf{OS.}  Network stacks (TCP/IP, TLS) provided reliable, secure
transport.  The OS enforced packet-level policies.

\textbf{Cloud-OS.}  The Istio/Envoy service mesh moved cross-cutting concerns
--- mTLS, circuit-breaking, retries, rate limiting, distributed tracing ---
out of application code into a transparent sidecar proxy.  Envoy routes on
HTTP/2 headers and gRPC metadata; its L7 processing model was designed for
structured, bounded request payloads.

\textbf{Semantic gap.}  Envoy is blind to prompt content, tool-call semantics,
and inferencing payloads.  Decisions are made based on static fields in request header, whereas content relevant for processing AI requests may appear in request body whose processing may require statefull, non-linear, re-entrant pipelines. Also,
MCP/uTCP/A2A evolve faster than Envoy release cycles.

\textbf{{\AOS} primitive.}  Alternative light-weight implementations (Rust-based) emerge for AI network infrastructure with flexible request processing pipelines~\cite{ibm2026praxis, agentgateway}. They can be transparently integrated with agents using a sidecar pattern to intercept all outbound tool calls,
validate SPIFFE identity, enforce authorization, exchange tokens (RFC~8693), and emit semantic telemetry.  MCP and agent
gateways provide a single point of access, routing, and access control.

% ── XVII. rossoctl: A Prototype {\AOS} ──────────────────────────────────────────
\section{Open source prototypes of {\AOS} primitives}
\label{sec:rossoctl}

While the emergence of a complete {\AOS} is a multi-year endeavor, the
thirteen primitives described above are not purely theoretical.  We have
implemented a subset of them in \textit{rossoctl} \cite{rossoctl2025github} ---
an open-source {\AOS} prototype available at
\url{https://github.com/rossoctl/rossoctl} --- built on Kubernetes. As such, rossoctl currently targets cloud and datacenter deployments; personal or edge agents are outside its scope.

\begin{figure}[htbp]
  \centering
  \includegraphics[width=\linewidth]{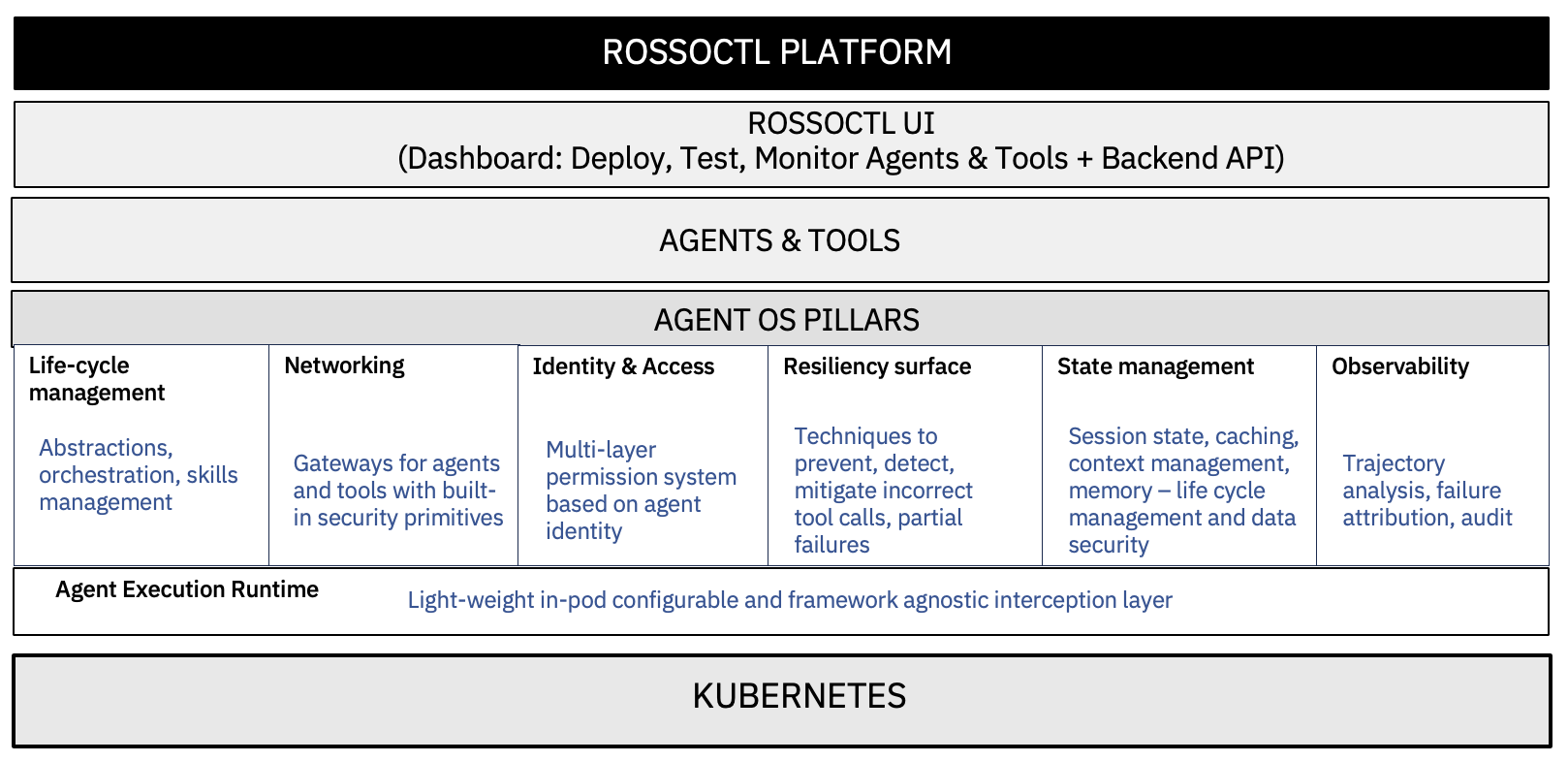}
  \caption{The rossoctl architecture, implementing several {\AOS}
           primitives on Kubernetes.  Platform pillars implement
           lifecycle orchestration, networking (MCP Gateway, service mesh),
           security (AuthBridge, SPIFFE/SPIRE, OAuth/OIDC), and observability
           (tracing, network visualization).  Agent workloads (A2A-compatible
           frameworks run above the platform layer.}
  \label{fig:rossoctl}
\end{figure}

The rossoctl architecture (Figure~\ref{fig:rossoctl}) embodies several design principles derived from the
OS analogy.  \textbf{Platform-enforced primitives}: authorization (AuthBridge),
identity attestation (SPIFFE/SPIRE), and network policy (MCP Gateway, Gateway
API) are platform services that agents consume transparently via sidecars and
CRDs --- they cannot be bypassed by application code, analogous to kernel-level
enforcement in classical OS security.  \textbf{Kubernetes-native control plane}:
rossoctl extends Kubernetes rather than replacing it, using Custom Resource
Definitions (\texttt{MCPServer}, \texttt{VirtualMCPServer}) and the Gateway API
to express agent platform policy in the same language as cloud infrastructure
policy.  This mirrors the cloud OS's strategy of extending, not discarding,
its predecessor. \textbf{Framework agnosticism}: Agent Execution Environment implemented in the form of in-network side-cars provides a transparent point of isolation, observability, and control. Thanks to it, rossoctl supports
LangGraph, CrewAI, Marvin, AutoGen, and generalist harnesses through a
common sidecar interface, just as a classical OS runs programs regardless of
the language they were written in.  \textbf{Composable new components}:
AuthBridge, Agent Execution Environment and the MCP Gateway are new components
designed specifically for agentic primitives that have no adequate
Kubernetes-native equivalent. 

%The most credible signal of {\AOS} feasibility we have found is this:
%rossoctl-hosted agents are used by our engineering team for daily work --- code
%review, issue triage, documentation, test generation.  When the platform
%builders themselves trust the platform in production, that constitutes stronger
%evidence of readiness than any benchmark.  We track three success dimensions:
%\textit{technical} (loop detection rate, token reduction, task success on
%benchmarks), \textit{adoption} (internal and external production use, client
%feedback), and \textit{ecosystem} (open source community growth, academic
%engagement).

\textbf{Other open-source {\AOS} efforts.}  rossoctl is not alone in
exploring what an {\AOS} should look like.  Several open-source projects instantiate individual Agent-OS primitives as a byproduct of their primary focus on agent development frameworks.
\textit{Microsoft AutoGen} \cite{autogen2023} provides a multi-agent
conversation framework that addresses orchestration, agent lifecycle, and
inter-agent communication, with an active open-source community building toward
production-grade runtime guarantees.  \textit{LangChain / LangGraph}
\cite{langchain2023} offers stateful, graph-based agent orchestration together
with an ecosystem of tool integrations and memory backends, representing one of
the most widely deployed bases for production agent applications.
\textit{OpenAI Swarm} \cite{openai_swarm2024} is an experimental,
minimalist multi-agent orchestration framework that explores lightweight
handoffs and context passing between agents. Kagent~\cite{kagent2026} experiments with orchestrating agents on Kubernetes platform, and Openshell~\cite{openshell2026} provides secure sandbox management for agent harnesses. Each of these projects
instantiates a different subset of the thirteen {\AOS} primitives described
in this paper; none yet provides the full platform layer --- with enforced
identity, authorization, behavioral guardrails, and semantic observability ---
that a mature {\AOS} requires.  The convergence of these efforts, and the
emergence of shared standards they can build to, is precisely the Phase~2
consolidation the field needs.

% ── XVIII. Discussion: What the Prior Waves Teach Us ─────────────────────────
\section{Discussion: What the Prior Waves Teach Us}
\label{sec:discussion}

The prior OS eras offer a methodology, not just vocabulary.  Four lessons are
directly applicable to the agentic wave.

\textbf{Lesson 1: Standardization is the exit from experimentation.}
Both the OS and Cloud-OS eras went through the same three-phase arc described in
Section~\ref{sec:intro}: experimentation, articulation, consolidation.  The
transition from Phase~1 to Phase~2 was not organic --- it required deliberate
community effort to agree on semantics.  POSIX standardized what a ``process''
means, what ``virtual memory'' guarantees, and what a ``system call'' can and
cannot do.  Without POSIX, every Unix application would have been tied to a
single vendor's implementation; with it, a generation of portable software
became possible.  Kubernetes standardized what a ``pod'' means, what a ``service''
guarantees, and what a ``health check'' requires.  Without it, the cloud-native
ecosystem would have remained a fragmented set of incompatible orchestrators.
The agentic community is currently producing the equivalent of pre-POSIX Unix
variants and pre-Kubernetes orchestrators: useful, innovative, but non-composable.
The thirteen primitives in this paper are a proposal for the agentic
equivalent of POSIX --- the abstract interface that portable agentic
applications should be written to, regardless of which vendor's platform runs
beneath them.  Their value is not in any single implementation but in the
stability and portability that agreed semantics enable.

\textbf{Lesson 2: Build on, not in replacement of, the prior layer.}
The Cloud-OS did not replace the OS; it ran on top of it.  Kubernetes pods
run inside Linux namespaces; Kubernetes RBAC extends POSIX UIDs to cluster
resources; service mesh mutual TLS extends kernel-level authentication to
service-to-service communication; PersistentVolumes extend the POSIX file
system to distributed storage.  Every Cloud-OS innovation extended an OS
abstraction to a new substrate.  The {\AOS} must follow the same pattern: agent
containers run inside Kubernetes pods; agent authorization builds on
SPIFFE/SPIRE, which builds on Kubernetes ServiceAccounts, which build on POSIX
UIDs; agent memory builds on existing database consistency models; AI-aware
networking builds on the Cloud-OS service mesh sidecar pattern.  Every {\AOS}
primitive in Table~\ref{tab:primitives} has both an OS and a Cloud-OS origin
precisely because inheriting proven semantics is cheaper, safer, and more
interoperable than reinventing them.

\textbf{Lesson 3: Specify only what is genuinely new.}  The Cloud-OS community
extended OS semantics precisely where the new substrate --- network instability,
node failure, dynamic scaling --- broke the classical assumptions, and not
elsewhere.  The {\AOS} community must apply the same discipline.  The semantic
gap in Section~\ref{sec:defs} --- stochasticity, action reversibility,
goal-directed misbehavior --- is the exact boundary of what is genuinely new
in the agentic substrate.  Kubernetes namespaces still isolate agent containers;
POSIX file systems still store agent artifacts; OpenTelemetry spans still trace
service calls.  Specifying only the delta keeps the abstraction surface small
and makes standardization tractable.

\textbf{Lesson 4: Platform enforces what application code only requests.}
Both the OS and Cloud-OS eras established that application code cannot be trusted to enforce
resource limits, security boundaries, or correctness invariants.  The OS
enforces isolation via the MMU; Kubernetes enforces resource limits via
cgroups.  This conclusion is even more relevant when an application is an agent, arguably the most stochastic component introduced into the ecosystem. The {\AOS} must enforce authorization, guardrails, context limits, and
observability through infrastructure mechanisms, not agent code.  It must support resiliency, reliability, and protect the external systems from corruption caused by unreliable agents.

% ── XVIII. Open Research Agenda ──────────────────────────────────────────────
\section{Open Research Agenda}
\label{sec:agenda}

Table~\ref{tab:primitives} summarizes the solution status for each proposed
primitive, distinguishing where the classical analog provides adequate semantics
from where new specification work is required.  It also suggests a time horizon where these primitives are likely to converge. The eight problems below are
the semantic specification tasks that correspond to the open entries:

\begin{enumerate}
  \item \textbf{Stochastic scheduling theory.} A scheduling framework for
        workloads whose duration, resource consumption, and success probability
        are random variables requires new formal models.

  \item \textbf{Evaluation methodology for stochastic systems.} How do you test
        a system whose correct behavior is a probability distribution?
        Classical testing assumes determinism; chaos engineering for agents ---
        systematically injecting stale memory, tool failures, and context
        truncation --- has no established methodology \cite{cemri2025mast}.

  \item \textbf{Semantic skill interface contracts.} A formal specification
        for skill pre/post-conditions, side effects, and model
        compatibility --- analogous to typed function signatures but operating
        over natural language semantics \cite{chen2026skvm}.

  \item \textbf{Agent-aware information flow analysis.} Tracking how
        information propagates across agent boundaries and enforcing flow
        policies that reason about \textit{semantic content}, not just bytes
        \cite{verifycommit2026}.

  \item \textbf{Software supply-chain security for AI-generated code.} New
        frameworks for detecting hallucinated dependency names, verifying
        AI-generated code security, and generating SBOMs for codebases where
        the developer cannot attest to dependency origin
        %\cite{mendio2025hallucinated,cloudsmith2025supply}.
        \cite{mendio2025hallucinated}.

  \item \textbf{Formal policy languages for natural-language guardrails.} A
        policy language expressive enough to capture semantic behavioral
        constraints, efficiently evaluateable,
        and robust against adversarial manipulation
        \cite{apollo2024scheming}. Recent work has begun to address this problem using a DSL with formal trigger-predicate-action semantics~\cite{wang2026agentspec} and probabilistic model checking~\cite{probguard2025}; or deontic temporal logic with a formal substrate for obligations and prohibitions over agent behavior~\cite{deontic2025v2026deontictemporallogicformal}. What remains open is a policy language expressive enough to reason over natural language outputswith decidable or efficiently approximable evaluation, and robust against adversarial manipulation of the policy predicate itself.

  \item \textbf{Principled context eviction theory.} A formal framework for
        context window management with  guarantees analogous to
        Bélády's algorithm --- characterizing the optimal eviction policy for
        a given class of agent tasks \cite{packer2023memgpt}.

  \item \textbf{Agentic RL and meta-learning.}  Agents trained via
        reinforcement learning or meta-learning interact with the platform in
        qualitatively different ways from inference-only agents: they generate
        reward signals, update their behavior in response to environment
        feedback, and may transfer learned policies across tasks.  The {\AOS}
        must understand these interactions --- how scheduling, memory, and
        context management decisions affect what an RL agent learns, and how
        to provide safety guarantees for agents whose behavior is not fixed at
        deployment time but continues to evolve in production.  This is an area
        in flux; the platform primitives needed to support agentic RL safely at
        enterprise scale remain largely undefined \cite{karpathy2025llmwiki}.
\end{enumerate}

% ── XIX. Conclusion ───────────────────────────────────────────────────────────
\section{Conclusion}
\label{sec:conclusion}

Every major wave of platform software passes through the same arc:
experimentation, articulation, consolidation.  Pre-POSIX, Unix variants
proliferated and programs were not portable.  POSIX standardized the
abstractions and enabled a generation of portable software.  Pre-Kubernetes,
cloud orchestrators fragmented and workloads were not portable across clouds.
Kubernetes standardized the abstractions and enabled the cloud-native ecosystem.
The agentic wave is in its pre-POSIX moment.  Frameworks proliferate; agents
are not portable; platforms do not compose.  The path forward is the same one
the field has followed twice before: agree on abstractions, specify their
semantics, and consolidate.

We have proposed thirteen {\AOS} primitives as a candidate set of those
abstractions.  Each is derived from a proven OS abstraction, extended to handle
stochastic execution, natural-language mediation, and autonomous real-world
action.  Each carries an open specification problem --- the semantic work that
must be completed before the abstraction can be standardized and platforms can
portably implement it.  These problems are not a sign of immaturity; they are
the normal work of Phase~2.  

The empirical urgency is real: a wide-ranging industry study of AI deployments in enterprise found that 95\% of them fail, and these failures are attributed not to the quality of models but to brittle workflows, context gaps, governance, and integration fragility~\cite{challapally2025genai}.  The answer is not more frameworks. It is agreed abstractions, precise
semantics, and platform-level enforcement.   

\section{Acknowledgment}
The authors acknowledge the aid of AI Claude Sonnet 4.6 to assist in the drafting and structuring of this manuscript.
% ── References ────────────────────────────────────────────────────────────────
\bibliographystyle{IEEEtran}
\bibliography{references}

\end{document}